# Towards a Consistent, Sound and Complete Conceptual Knowledge

Gowri Shankar Ramaswamy[#1], F Sagayaraj Francis[#2]

[#1,2]*Department of Computer Science & Engineering*
*Pondicherry Engineering College, Puducherry, India*

*Abstract*— Knowledge is only good if it is sound, consistent and complete. The same holds true for conceptual knowledge, which holds knowledge about concepts and its association. Conceptual knowledge no matter what format they are represented in, must be consistent, sound and complete in order to realise its practical use. This paper discusses consistency, soundness and completeness in the ambit of conceptual knowledge and the need to consider these factors as fundamental to the development of conceptual knowledge.

## I. INTRODUCTION

Possessing knowledge at a conceptual level is akin to having a tangible record of the way human beings understand the world [1]. Studying various aspects of conceptual knowledge is very intriguing since challenges are plenty [2], some of them being the correctness of the conceptual knowledge per se. Any knowledge must be in some concrete form in order to be processed and exchanged. In computer science or information science, knowledge representation is a whole field unto itself as the applicability of knowledge rests primarily on its representation [3] [4]. Irrespective of the way of representation, the question of correctness always persists. Correctness, per se, depends on three properties namely consistency, soundness and completeness [5] [6]. Understanding what these properties mean is the initial step towards devising conceptual knowledge that is consistent, sound and complete. These properties check that the foundation upon which conceptual knowledge is built is strong. Conceptual knowledge is going to be consumed by computing systems to improve themselves. This knowledge is not intended to replace anything that already exists in the computing infrastructure. It is just going to augment the existing resources, add value to what already exists and what is already being offered. Hence the usefulness of conceptual knowledge is very much reliant on its own strength which comes from having a solid foundation. Conceptual knowledge, as propounded in literature, is built over a mathematical foundation, typically using linear algebraic vector structures. Gärdenfors [4] [8] conceptualized a conceptual space in which every concept is defined based on a set of attributes or quality dimensions as a linear combination of vectors. The primitive design had limitations that led to further research in this field [2] [9]. Every improvement in this field aims only to make the knowledge comprehensive and useful but in this quest it is very important to ensure that the knowledge is consistent, sound and complete. This paper elucidates these criteria in the context of conceptual knowledge. In section II a brief background on conceptual knowledge is discussed to provide clarity on the technical premise of this paper. Section III, the crux of the paper, elaborates the meaning of consistency, soundness and completeness in conceptual knowledge and justifies the need to meet these considerations. Section IV gives the concluding remarks on the paper with guidance for future.

## II. CONCEPTUAL KNOWLEDGE

Conceptual knowledge built over conceptual space of concepts is a linear combination of vectors, each vector representing a quality dimension. A quality dimension is an attribute or a characteristic feature of a given concept. Quality dimensions are assumed to be linearly independent but that need not be the case always [2] [10]. Interdependence between quality dimensions has a covariant impact on the quality dimensions. Nevertheless, for the sake of simplicity, dimensions are assumed to be independent.

In the conceptual vector space, quality dimensions can be scalar or vector or a set of vectors [11]. The conceptual space therefore spans over a number of domain and subdomains.

Conceptual vector space of concepts $C_n$ is defined as

$$C_n = \{(q_1, q_2, ..., q_n) \mid q_i \in C\} \quad (1)$$

where $q_i$ are the quality dimensions.

If a quality dimension represents a domain, then

$$q_i = S_n = \{(s_1, s_2, ..., s_n) \mid s_k \in S\} \quad (2)$$

Similarly, all concepts and their quality dimensions can be iteratively defined.

### III. CONSISTENCY, SOUNDNESS, COMPLETENESS OF CONCEPTUAL KNOWLEDGE

Conceptual knowledge, as stated earlier, comprises of concepts and connections between concepts, usually confined to a specific domain. The domain narrows the meaning of concepts to a particular subject of discussion since a concept can exist in multiple domains and can have multiple meanings or interpretations [11]. The correctness of conceptual knowledge ultimately trickles down to the extent to which it is able to replicate human understanding. So the efficiency of conceptual knowledge can be tested by obtaining feedback from end users. However, this is only possible after the knowledge has been designed and deployed for practical use. Since the design of the knowledge follows a formal foundation, it is possible to ascertain its correctness by setting some standards and testing the compliance of the design of the knowledge to this standard. As conceptual knowledge relies heavily on the definition of the concepts, it is worth noting here that, in mathematics, definitions can be either intensional or extensional [12]. Intensional definition is the process of giving meaning by specifying all the properties required to come to that definition i.e. all the necessary and sufficient properties that are to be possessed in order to qualify to belong to a particular group. On the contrary, extensional definition is an enumerative process wherein the qualifying members are enumerated where the properties necessary for qualification cannot be clearly listed as in the case of intensional definition.

The definition of concepts discussed in Section II is an intensional definition. Since conceptual knowledge is a collection of concepts and since this collection or set has every chance of being an infinite set, an extensional definition of concepts would be impossible. Conceptual knowledge, in order to gain acceptance, must have an intensional definition that satisfies the criteria of consistency, soundness and completeness. The reasons for this necessity are discussed in the subsequent paragraphs.

#### A. Consistency

Consistency implies that the definition does not allow contradictions. In the realm of conceptual knowledge, this means that a concept should not possess contradicting properties. It could also imply that concepts must not be ambiguous in its definition. This property ensures that the knowledge is well defined and doesn't contain paradoxes which may affect the efficiency of the working of the knowledge.

#### B. Soundness

Soundness means that nothing invalid can be derived using the intensional definition. This means that the definition is sound if and only if it is valid and all the premises are actually true. The definition is valid if the premises that lead to the conclusion are logical and can be verified. Premises of the definition of concept are the quality dimensions or attributes mentioned in Section II that actually describe a concept. The design of conceptual knowledge must be in such a way that the intensional definition leads to a well defined concept and this makes the knowledge sound.

#### C. Completeness

Completeness means the reverse of soundness i.e. all valid concepts must be derivable from the intensional definition. This ensures the fact that the knowledge is designed to accommodate only valid concepts and if a concept is said to be valid, it should have a corresponding intensional definition with properties describing them.

After having clarified the definitions of consistency, soundness and completeness, it is worth noting that some of the core design elements can have an impact on these factors. Consider, for example, that a concept with certain properties and the same concept with contradictory properties exist in the knowledge. This could seem as a violation in the consistency of the knowledge. But, say, if the concept had a temporal or a contextual dimension to it that gave rise to a copy of the same concept but with a modified definition. If this is the case, both the conceptual definitions are valid provided there is some element in the design that distinctly distinguishes them based on the temporal or contextual dimension. Building the knowledge with such design considerations contributes to ensuring its consistency, soundness and completeness.

IV. CONCLUSIONS

The thoughts posited in this paper are meant to augment the need for a comprehensive conceptual knowledge which is an area of active research. Having a strong theoretical foundation not only gives stability to conceptual knowledge but also helps to steer research in the field in a disciplined path. These criteria can also be used as guidelines whenever improvements are made to the design of conceptual knowledge so that integrity is maintained.